\documentclass[journal=jacsat,manuscript=article]{achemso}

\usepackage{chemformula} 
\usepackage[T1]{fontenc} 

\author{Yang SHAO}
\email{yang.shao.kn@hitachi.com}
\affiliation[hitachi]
{Research and Development Group, Hitachi, Ltd.}
\author{Toshie YAGUCHI}
\affiliation[hightech]
{Hitachi High-Tech Corporation}
\author{Toshiaki TANIGAKI}
\affiliation[hitachi]
{Research and Development Group, Hitachi, Ltd.}

\title[An \textsf{achemso} demo]
  {Denoising Low-dose Images Using \\ Deep Learning of Time Series Images}

\abbreviations{IR,NMR,UV}

\begin{document}







\begin{abstract}
Digital image devices have been widely applied in many fields, including scientific imaging, recognition of individuals, and remote sensing. As the application of these imaging technologies to autonomous driving and measurement, image noise generated when observation cannot be performed with a sufficient dose has become a major problem. Machine learning denoise technology is expected to be the solver of this problem, but there are the following problems. Here we report, artifacts generated by machine learning denoise in ultra-low dose observation using an in-situ observation video of an electron microscope as an example. And as a method to solve this problem, we propose a method to decompose a time series image into a 2D image of the spatial axis and time to perform machine learning denoise. Our method opens new avenues accurate and stable reconstruction of continuous high-resolution images from low-dose imaging in science, industry, and life. 
\end{abstract}

\section{Introduction}

In the realm of scientific research, Electron Microscopy has emerged as an indispensable tool, offering unprecedented insights into the microscopic world. Despite its profound contributions, capturing high-quality images under low-dose conditions remains a formidable challenge. This challenge stems from the inherent trade-off between image quality and exposure time: while longer exposure times can yield high-resolution images, they also increase the risk of sample damage \cite{egerton2013control}. Conversely, shorter exposure times, although safer for the sample, often result in images plagued by noise and low resolution \cite{lee2014electron}. There are several attempts to denoise low-dose images by using machine learning \cite{lehtinen2018noise2noise, krull2019noise2void, li2023real}. However, the denoising for live movies and time series images are still challenging. 

In light of these challenges, this paper proposes a novel approach that leverages the power of machine learning and three-dimensional (3D) synthesis to enhance low-dose microscopy images. At the heart of our method is a two-step process: the generation of synthetic data "(Section 2.2)" followed by microscope calibration "(Section 2.)". The synthetic data, generated using a simulator, incorporates various modes of random noise, providing a diverse and robust training set for our base model. The microscope calibration, performed prior to actual microscopy observations, involves fine-tuning the base model with low-noise high-resolution images acquired using intense irradiation. 

In addition to this machine learning-based enhancement, we introduce a computationally efficient 3D image processing technique based on three-directional synthesis. This technique begins by capturing a three-dimensional tensor representing the observed scene. The tensor is then sliced along three orthogonal directions, yielding three continuous image segments. Each segment undergoes individual noise reduction enhancement processes, resulting in three separate 3D tensors. The final image is obtained by computing the average of these tensors, thereby creating a low-noise high-resolution representation of the original scene. 

By combining details from multiple perspectives, our method ensures accurate and stable reconstruction of continuous high-resolution images from low-dose observations. Note that this method can be applied to time-resolved electron microscopy \cite{kuwahara2022transient}, which also requires low-dose imaging. This research opens new avenues for high-resolution microscopy in challenging lighting conditions. The potential applications of this work span across various fields, including materials science, biology, and medical diagnostics. 

\section{Research Method}

\subsection{Pre-training fine-tuning paradigm}

In the early days of machine learning, feature engineering and handcrafted kernel functions were the norm. Techniques like Support Vector Machines (SVMs) thrived on these manual designs. However, the landscape shifted dramatically with the rise of deep neural networks. Features no longer demanded human craftsmanship; instead, the paradigm shifted toward designing neural network structures tailored to specific problems, employing diverse training algorithms. Today, machine learning is facing another paradigm shift, and a new era has emerged - the era of universal basic models. This paradigm dispenses with the need for bespoke neural network architectures for every distinct problem. Instead, it harnesses the power of pre-trained base models, followed by fine-tuning.  

The main two phases of pre-training fine-tuning paradigm are as follows. Pre-training: Constructing a versatile scaffold—a base model—by training it on a wealth of diverse tasks. This base model learns fundamental features, akin to a polymath absorbing knowledge across disciplines. It becomes a repository of general patterns, ready to adapt. Fine-Tuning: Faced with a specific problem, instead of reinventing the wheel, we fine-tune the base model to adapt it to the nuances of the task, becoming a specialized tool.  

The major advantages and benefits are as follows. Transfer Learning: Leveraging pre-trained base models accelerates learning. We inherit wisdom from a multitude of tasks, akin to standing on the shoulders of giants. Customization: Fine-tuning tailors the base model to specific challenges. It’s like adjusting a telescope’s focus for celestial clarity. No need to start from scratch, we build upon existing knowledge. Efficiency: Gone are the days of designing bespoke architectures and conducting separate training sessions. Base models provide a head start, and fine-tuning ensures adaptability. 

In summary, the pre-training fine-tuning paradigm bridges the gap between the massive amounts of data required by traditional machine learning methods and the limited data available for domain-specific needs. 

\subsection{Three-dimensional (3D) synthesis}

In recent years, video denoising has gained significant attention due to its applications in various domains such as surveillance, entertainment, and medical imaging. Traditional methods often rely on handcrafted features or filters, which may not fully exploit the rich spatial and temporal information present in videos. Therefore, deep learning techniques have emerged as powerful tools for denoising tasks. However, when applying deep neural networks to video denoising, a common approach involves using three-dimensional tensors (3D volumes) as input. These 3D tensors represent video frames along the spatial and temporal dimensions. While effective, this approach has limitations. The resulting neural network can become quite large, requiring substantial computational resources for both training and inference.

To address this challenge, we propose a method that leverages sliced three-dimensional tensors. Here’s how it works:

\begin{itemize}
\item Tensor Slicing:
Given the original 3D tensor representing a video sequence, we slice it along three orthogonal directions (X, Y, and T). This results in three stacks of 2D images, each capturing information from a specific perspective.
The X-slice contains frames along the horizontal spatial dimension.
The Y-slice contains frames along the vertical spatial dimension.
The T-slice contains frames along the temporal dimension.
\item Three Separate Models:
Next, we build three separate neural network models, each learned to denoise the 2D images from one of the slices. These models are smaller and more manageable than a single large network.
The X-model processes the X-slice images.
The Y-model processes the Y-slice images.
The T-model processes the T-slice images.
\item Fusion and Synthesis:
After denoising each slice independently, we combine the results. Specifically:
The denoised X-slice, Y-slice, and T-slice are reassembled into a 3D tensor.
The final denoised video is synthesized from this reassembled tensor.
\end{itemize}

By breaking down the problem into three smaller models, our approach reduces the overall complexity of the denoising process. Training becomes more efficient, as each model focuses on a specific aspect of denoising. During inference, the computational speed improves due to the smaller model sizes.
The schematic diagram of three-dimensional slices is shown in Figure 1.

\begin{figure}[t]
  \centering
   \includegraphics[width=0.8\linewidth]{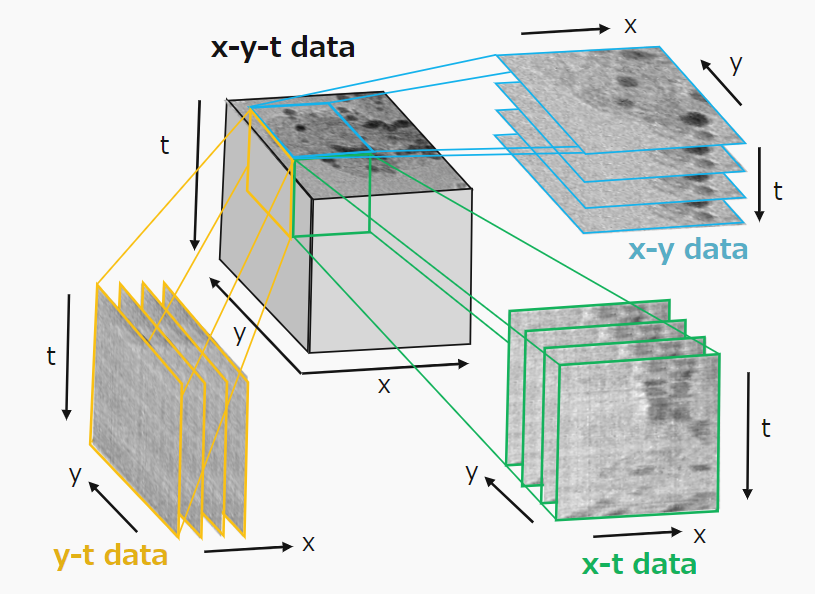}
   \caption{Schematic diagram of three-dimensional slices}
   \label{fig:xyt}
\end{figure}

\subsection{Artificial data for base model}

We initially used an electromagnetic field generated near the PN junction through simulation as our ground truth. After achieving satisfactory results, we further utilized actual electron microscope images of catalysts obtained from previous captures. To strike a balance between higher observation clarity and computational time requirements, we temporarily set an observation window size of 128x128. As computational power increases, we will be able to train models capable of processing higher-resolution images within the same time frame. Additionally, we can achieve processing of higher-resolution images without altering the model by segmenting the image and processing multiple observation windows, subsequently reassembling the results. 

To generate artificial data for training our foundational model, we introduced varying levels of different types of noise to the ground truth images. As two typical examples, we added Poisson noise and Gaussian noise. In order to enhance the foundational model’s ability to handle different types of noise, we will continue to incorporate additional noise types and explore various hybrid noise models. An example of actual captured images and images with artificial noise is shown in Figure 2. 

\begin{figure}[t]
  \centering
   \includegraphics[width=0.4\linewidth]{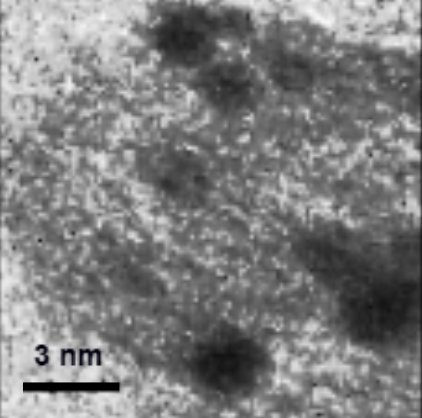}
   \includegraphics[width=0.4\linewidth]{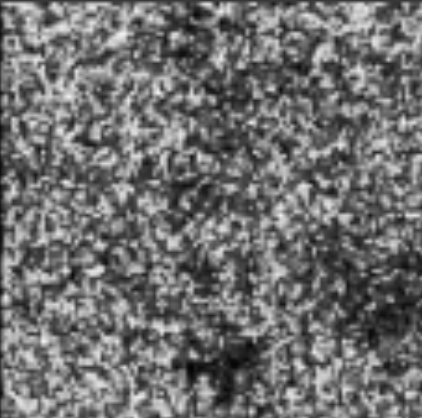}
   \caption{Actual captured images and images with artificial noise}
   \label{fig:data}
\end{figure}

\subsection{Structure of base model}

In the field of image generation, there are currently two prominent architectural approaches: autoencoder-based and Diffusion-based architectures. As an initial exploration, our research employed the well-known U-Net architecture. This architecture is specifically designed for semantic segmentation tasks. Let's delve into its components in detail:

\begin{enumerate}
    \item Contracting Path (Encoder):
        \begin{itemize}
            \item The contracting path follows the typical architecture of a convolutional network.
            \item It consists of repeated applications of two 3x3 convolutions (unpadded convolutions), each followed by a rectified linear unit (ReLU).
            \item Additionally, a 2x2 max pooling operation with a stride of 2 is used for downsampling.
            \item At each downsampling step, the number of feature channels is doubled.
        \end{itemize}
    \item Expansive Path (Decoder):
        \begin{itemize}
            \item The expansive path aims to recover the spatial resolution lost during downsampling.
            \item It consists of:
                \begin{itemize}
                    \item Upsampling of the feature map.
                    \item A 2x2 convolution ("up-convolution") that halves the number of feature channels.
                    \item Concatenation with the correspondingly cropped feature map from the contracting path.
                    \item Two 3x3 convolutions, each followed by a ReLU.
                \end{itemize}
            \item The cropping step is necessary due to the loss of border pixels in every convolution.
        \end{itemize}
    \item Final Layer:
        \begin{itemize}
            \item At the final layer, a 1x1 convolution is used to map each 64-component feature vector to the desired number of classes.
        \end{itemize}
\end{enumerate}

In total, the U-Net network comprises 23 convolutional layers. While this architecture served as our initial choice, we plan to explore other more efficient architectures in our subsequent research.

\subsection{Training of base model}

In our study, we used a dataset comprising 300 sets of 128 high-resolution images of size 128x128 pixels. These images were artificially corrupted with noise to serve as the input to our U-Net model as described above. The output of the U-Net model, which are the denoised images, were then compared with the corresponding noise-free high-resolution images. The Mean Squared Error (MSE) between the output and the noise-free images was computed and used as the loss function for training the U-Net model. The MSE loss function is mathematically represented as:

\[
\text{{MSE}} = \frac{1}{n} \sum_{i=1}^{n} (Y_i - \hat{Y}_i)^2
\]

where $Y_i$ represents the value of pixels in actual noise-free images, $\hat{Y}_i$ represents the value of pixels in denoised images produced by the U-Net model, and $n$ is the total number of pixels. In order to improve the learning effect, we normalized all pixels so that the values of all pixels fall between 0-1.

To optimize the U-Net model, we employed the Backpropagation algorithm, which is a widely-used method in training neural networks. Backpropagation computes the gradient of the loss function with respect to the model parameters and uses this gradient to update the parameters. The mathematical representation of the parameter update rule in Backpropagation is:

\[
\theta = \theta - \eta \cdot \nabla_\theta J(\theta)
\]

where $\theta$ represents the parameters of the model, $\eta$ is the learning rate, and $\nabla_\theta J(\theta)$ is the gradient of the loss function $J(\theta)$ with respect to the parameters. 

For the optimization process, we used the Adam optimizer. Adam, short for Adaptive Moment Estimation, is an optimization algorithm that computes adaptive learning rates for each parameter. In contrast to classical stochastic gradient descent, Adam keeps an exponentially decaying average of past gradients and past squared gradients, and uses these to scale the learning rate. We set initial learning rate of Adam as 0.001. 

In an effort to balance learning accuracy and efficiency, we employed PyTorch's autocast feature in our study. Autocasting is a feature that automatically chooses the data type - float32 or float16 - for each operation in order to improve the performance of the model while maintaining the accuracy. By using autocast, we were able to convert some of the float32 data to float16 for computation. This allowed us to leverage the benefits of float16, such as reduced memory usage and increased computational speed, without significantly compromising the accuracy of our model. This approach is particularly beneficial when training large models or when memory capacity is a limiting factor.

Due to the limitation of computing device memory space, we adopted a relatively small batch size of 4 and conducted learning for 50 epochs.

We trained three different U-Net models for the aforementioned three directions, x-y, x-t, y-t.

\subsection{Comparisons}

In our study, we employed a comparative approach to evaluate the effectiveness of our proposed denoising method. To this end, we utilized three well-established denoising techniques as benchmarks: fastNlMeansDenoising, GaussianBlur and bilateralFilter.

\begin{enumerate}
\item fastNlMeansDenoising: This non-local means denoising algorithm works by replacing the value of a pixel by an average of a selection of other pixels values. The selection is based on a patch similarity criterion, leading to effective noise reduction while preserving the structural details of the image.

\item GaussianBlur: Gaussian blur is a widely used effect in graphics software, typically to reduce image noise and detail. The visual effect of this blurring technique is a smooth blur resembling that of viewing the image through a translucent screen.

\item bilateralFilter: The bilateral filter is a non-linear, edge-preserving, and noise-reducing smoothing filter for images. It replaces the intensity of each pixel with a weighted average of intensity values from nearby pixels. This weight can be based on a Gaussian distribution, thus preserving sharp edges by systematically looping through each pixel and adjusting weights to the adjacent pixels accordingly.
\end{enumerate}

We used the opencv implementation of these methods. They were applied to the same noisy datasets for a fair comparison. The results of our method were then juxtaposed with the outcomes of these techniques, providing a comprehensive evaluation of our proposed method's performance in image denoising tasks. 

\section{Experiments}

We conducted two experiments. One was designed to test the learning performance of U-Net, and the other was aimed at evaluating the effectiveness of 3D synthesis. 

\subsection{Comparative Analysis of Denoising Techniques}

For the first experiment, we utilized electromagnetic field images near the PN junction generated by a simulator, as well as actual electron microscope images of catalysts that consist of platinum and carbon particles as our experimental data. The results of U-Net were compared with those of fastNlMeans, GaussianBlur, and bilateralFilter. We used the Mean Squared Error (MSE) of all pixels as the comparison metric. The PN junction data we used to consist of 5120 pairs of images with artificially added noise and their corresponding images before noise addition. We used four-fifths of this data for training and the remaining one-fifth for testing. The real image dataset captured by the electron microscope consisted of 300 groups of 128 consecutive images, each of size 128x128. We used five-sixths of this data for training and the remaining one-sixth for testing. 

The results are shown in Table 1. The table presents the Normalized Mean Squared Error (MSE) of pixels for different methods applied to various datasets. The MSE is measured in units of $10^{-3}$. For the PN junction in the x-y plane, the original noised data had an MSE of 175.08. After applying our U-Net, the MSE was significantly reduced to 8.91, which is the lowest among all methods. Other methods such as fastNlMeansDenoising, GaussianBlur, and bilateralFilter also reduced the MSE, but not as effectively as our U-Net. Similarly, for the PN junction in the x-t plane, the original noised data had an MSE of 156.12. Our U-Net was able to reduce this to 8.01, again achieving the best performance compared to other methods. For the catalyst particles in the x-y plane, the original noised data had a much higher MSE of 268.98. Our U-Net reduced this to 164.25, outperforming GaussianBlur, which resulted in an MSE of 176.33. The fastNlMeansDenoising and bilateralFilter methods were not applied in this case. These results demonstrate the superior performance of our U-Net in reducing the MSE in different scenarios, outperforming other commonly used methods. This suggests that our U-Net could be a valuable tool for noise reduction in these applications. 

\begin{table*}[h]
\centering
\begin{tabular}{|c|c|c|c|c|c|}
\hline
 & \textbf{Original noised} & \textbf{Our U-Net} & \textbf{fN} & \textbf{GB} & \textbf{bF} \\ \hline
\textbf{PN junction x-y} & 175.08 & 8.91 & 10.33 & 9.24 & 10.11 \\ \hline
\textbf{PN junction x-t} & 156.12 & 8.01 & 9.48 & 8.56 & 9.91 \\ \hline
\textbf{Catalyst x-y} & 268.98 & 164.25 & - & 176.33 & - \\ \hline
\end{tabular}
\caption{Normalized MSE of pixels (10e-3)}
\label{tab:my_label}
fN: fastNlMeansDenoising
GB: GaussianBlur
bF: bilateralFilter
\end{table*}

\subsection{Comparative Analysis of 3D synthesis}

\begin{figure}[t]
  \centering
   \includegraphics[width=1.0\linewidth]{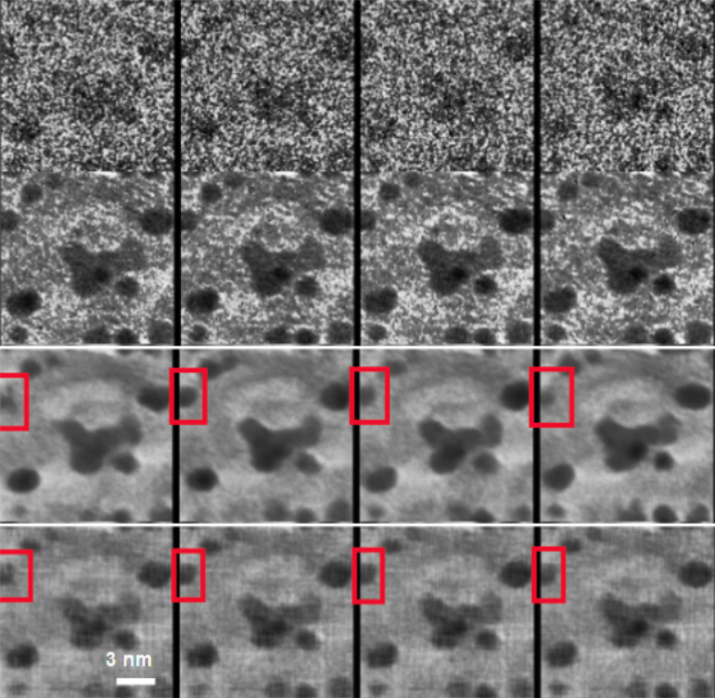}
   \caption{The particles in the red box maintain the continuity and stability of the display in 4 consecutive frames. (Line 1: Noisy in-put. Line 2: Actual observed image before adding noise. Line 3: Denoising result using only the x-y direction model. Line 4: Denoising results after 3D synthesis)}
   \label{fig:Effect}
\end{figure}

\begin{figure}[t]
  \centering
   \includegraphics[width=1.0\linewidth]{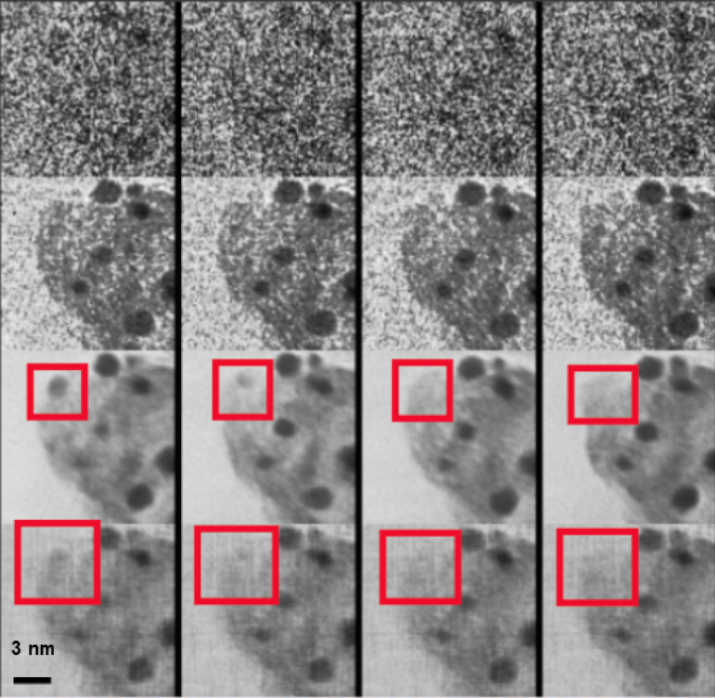}
   \caption{Phantom particles appeared in the first frame of the denoising result using only the x-y direction model. (Line 1: Noisy input. Line 2: Actual observed image before adding noise. Line 3: Denoising result using only the x-y direction model. Line 4: Denoising results after 3D synthesis.)}
   \label{fig:Effect2}
\end{figure}

For the second experiment, we utilized a real image dataset composed of 300 groups of 128 consecutive images, each of size 128x128, captured by an electron microscope. Each group of 128 images was sliced along the t, y, and x axes, resulting in 38,400 images each for the x-y, x-t, and y-t directions. We used five-sixths of these images for training and the remaining one-sixth for testing. 

We trained three separate U-Net models to handle slices from the x-y, x-t, and y-t directions. During testing, we reassembled the 128 slices from the same group into a three-dimensional tensor. The final synthesized tensor was obtained by averaging the three-dimensional tensors derived from slices in the three different directions. 

To validate the effectiveness of our 3D synthesis method, we compared the results of the final synthesized tensor in the x-y direction with the results obtained by directly applying the x-y U-Net model to the x-y images. This comparison allowed us to assess the performance and potential advantages of our 3D synthesis approach. 

Although the three-dimensional synthesis method has not yet shown significant effects in reducing the overall signal-to-noise ratio, we have observed two significant qualitative effects from the experimental results. Firstly, our observations demonstrated increased continuity and stability. Specifically, instances where particles temporarily faded or even disappeared due to noise interference were significantly reduced. This improvement suggests that the three-dimensional synthesis method enhances the reliability of our observations by mitigating the impact of noise. Secondly, we observed a significant decrease in the occurrence of phantom particles, which are artifacts that appear due to the aggregation of noise at specific locations. This reduction in phantom particles further attests to the effectiveness of the three-dimensional synthesis method in improving the quality of our observations. Figures 3 and Figure 4 provide examples of these improvements in continuity and stability, and the reduction of phantom particles, respectively. For figure 3, in the first and fourth frames of the denoising result using only the x-y direction model, the particles appear faded compared to the second and third frames. However, in the noising results after 3D synthesis, all particles maintain a relatively consistent clarity. For figure 4, in the first frame of the denoising result using only the x-y direction model, phantom particles appeared. However, this phenomenon did not occur in any of the noising results after 3D synthesis. These findings underscore the potential of the three-dimensional synthesis method in enhancing the accuracy and reliability of observations in our field of study. 

\section{Conclusion}

This paper has presented a preliminary exploration into the potential of enhancing low-dose microscopic images by harnessing the power of machine learning and three-dimensional (3D) synthesis. Our method, while novel, is just an initial step in addressing the complex challenge of improving image quality under low-dose conditions. 

Through two carefully designed experiments, we have demonstrated the potential of our approach. The first experiment, conducted on both simulated PN junction electromagnetic field images and real-world images of the platinum and carbon particles, provided encouraging results. Our U-Net-based denoising method showed promising performance compared to traditional techniques. The second experiment highlighted the benefits of our three-dimensional synthesis method. By applying this method to the actual electron microscope images of catalysts consisting of platinum and carbon particles, we observed a noticeable improvement in image continuity and stability.  

Looking forward, we see several areas for future exploration. We plan to investigate more sophisticated machine learning models and noise reduction techniques to further enhance the performance of our method. We also intend to optimize our 3D synthesis method further to improve computational efficiency and image quality. 

\bibliography{achemso-demo}

\end{document}